\pdfoutput=1
\documentclass[12pt]{article}
\usepackage{amsfonts}
\usepackage{amsmath}
\usepackage{graphicx}
\usepackage{epstopdf}
\usepackage{array}
\newtheorem{theorem}{Theorem}
\newtheorem{acknowledgement}[theorem]{Acknowledgement}

\begin{document}

\title{The Artists who Forged Themselves: Detecting Creativity in\ Art}
\author{Milan Rajkovi\'{c} \\
Institute of Nuclear Sciences Vinca, University of Belgrade, Serbia \and Milo%
\v{s} Milovanovi\'{c} \\
Mathematical Institute of the Serbian Academy of Sciences and Arts, Belgrade}
\maketitle

\begin{abstract}
Creativity and the understanding of cognitive processes involved in the
creative process are relevant to all of human activities. Comprehension of
creativity in the arts is of special interest due to the involvement of many
scientific and non scientific disciplines. Using digital representation of
paintings, we show that creative process in painting art may be objectively
recognized within the mathematical framework of self organization, a process
characteristic of nonlinear dynamic systems and occurring in natural and
social sciences. Unlike the artist identification process or the recognition
of forgery, which presupposes the knowledge of the original work, our method
requires no prior knowledge on the originality of the work of art. The
original paintings are recognized as realizations of the creative process
which, in general, is shown to correspond to self-organization of texture
features which determine the aesthetic complexity of the painting. The
method consists of the wavelet based statistical digital image processing
and the measure of statistical complexity which represents the minimal
(average) information necessary for optimal prediction. The statistical
complexity is based on the properly defined causal states with optimal
predictive properties. Two different time concepts related to the works of
art are introduced: the internal time and the artistic time. The internal
time of the artwork is determined by the span of causal dependencies between
wavelet coefficients while the artistic time refers to the internal time
during which complexity increases where complexity refers to compositional,
aesthetic and structural arrangement of texture features. The method is
illustrated by recognizing the original paintings from the copies made by
the artists themselves, including the works of the famous surrealist painter
Ren\'{e} Magritte.

\end{abstract}

\section{Introduction}

Digital image analysis methods have advanced in the past decade at an
accelerated pace and the interdisciplinary interaction of scientists
involved in the formulation and application of these methods, on one side,
and art experts on the other, has opened up new possibilities for the
advancement of knowledge of interest to both groups. The impetus for such
advancement is certainly due to the availability of high resolution images
of rich colour representation, among other things. One of the most
interesting and intriguing problems related to the use of art image
processing tools and methods is the artist identification in the sense of
indisputable attribution of the artist to the work of art \cite{berezhnoy},
\cite{li}, \cite{lyu}, \cite{jafarpour}, \cite{johnson}. In order to achieve
this goal, experts often rely on a combination of technical data obtained by
the use of sophisticated equipment for mechanical, chemical and optical
inspection of the art works and the visual inspection by art scholars
supplemented by information provided by art historians. Recently, image
processing techniques have appeared which analyse the higher-level features
of the painting, such as texture and brush strokes using 2-dimensional
wavelet transform or its complex counterpart \cite{johnson}, a technique
which is relevant for our approach presented here. Although these techniques
are sophisticated and in the early stage of development, and in spite of
encouraging results, there are certain weaknesses that leave ample room for
improvement. In general, all image processing methods require the original
work of art or the training set of original paintings in order to make the
comparison with the works of doubtful origin or uncertain authorship.

Our approach is based on the premise that the creativity is a process of
artist's self-organization on the mental level reflected in the
self-organization of forms, patterns, textures and brush strokes of the
painting which determine the aesthetic quality of the artwork. Recognition
of creativity as self-organization has appeared a few times in the
literature, notably in \cite{nalimov}, \cite{zausner} and in a fascinating
book by Rudolph Arnheim \cite{arnheim}. Arnheim writes: "The actual
functioning of a painting or a piece of music is all mental, and the
artist's striving toward orderliness is guided by the perceptual pulls and
pushes he observes within the work while shaping it. To this extent, the
creative process can be described as self-regulatory. However, here again,
as in the physiological mechanism mentioned above, it is necessary to
distinguish between the balancing of forces in the perceptual field itself
and the "outside" control exerted by the artist's motives, plans and
preferences. He can be said to impose his structural theme upon the
perceptual organization. Only if the shaping of aesthetic objects is viewed
as a part of the larger process, namely the artist's coping with the tasks
of life by creating his works, can the whole of artistic creativity be
described as an instance of self-regulation". Arnheim wrote this work in
1971 under a strong influence of Gestalt psychology and before the concept
of self-organization was scientifically interpreted in the works of
Prigogine on far-from-equilibrium dynamical systems \cite{prigogine}. More
recently, Zausner has written: "Creating and viewing visual art are both
nonlinear experiences. Creating a work of art is an irreversible process
involving \textit{increasing levels of complexity} and unpredictable events"
(italics by the present authors). Increasing complexity in time is our
apprehension of self-organization and represents our main guiding principle
in the analysis and comparison of the works of art.

\section{Complexity, self-organization and the wavelet decomposition method}

The central concept in our framework \cite{we} is self-organization which is
a ubiquitous concept related to the organization and dynamics of complex
systems. In general, self-organization denotes a spontaneous emergence of
structures and organized behavior without any external influence in systems
consisting of a large number of interconnected elements. In general, due to
the feedback relations between constitutive components, the dynamics of
self-organizing systems is non-linear. Self-organization indicates a
spontaneous increase in structural entanglement (complexity) of a system over
time. Since there is no unique definition of complexity there are a number
of ways to characterize it depending on the context and scientific interest.
Our approach has been influenced by the method of computational mechanics,
developed by J. Crutchfield and his collaborators, which focuses on the
measure of organization in the systems and on qualitative and quantitative
description of structure and patterns \cite{crutch}, \cite{crutch2}.
According to this program the organization of a process is its causal
architecture embodied in the key concept, the $\epsilon $-machine, which
reveals the structure of connections between causal states in the temporal
domain. The (statistical) complexity of a process is defined as the minimal
information necessary for optimal prediction, according to the proposition
in \cite{grassberger}, where the term " the true measure of complexity" was
used. An operational and practical formalization of this definition, in our
framework, is based on the wavelet decomposition of the data with the causal
architecture embodied in the \textit{wavelet}-machine \cite{we}.

The wavelet transform in the one-dimensional case (1-D), decomposes the
signal in terms of the shifted (in space or in time) and dilated (scaled)
versions of a wavelet function, which can be considered as a motive or
template. The signal then represents a superposition of these wavelet
templates with appropriate weights, which are known as wavelet coefficients.
In two dimensions wavelets acquire an additional attribute of orientation,
namely horizontal, vertical or diagonal. Three additional orientations may
be generated by the complex wavelet transform. An image under consideration
may be represented as a superposition of wavelet templates on a grid with
appropriate coefficients. Figure 1 illustrates the 1-D wavelet and the 2-D
wavelet which consists of three wavelets, namely horizontal, vertical and
diagonal. In the dyadic representation of scale and time (or space), which
is the standard practice, each wavelet coefficient has 2 (1-D) or 4 (2-D)
successors on a finer scale forming a binary tree structure (1-D) or
quad-tree (2-D) respectively, represented in Fig. 2.

\begin{figure}
  \begin{center}
  \includegraphics[
   width=8cm
   ]{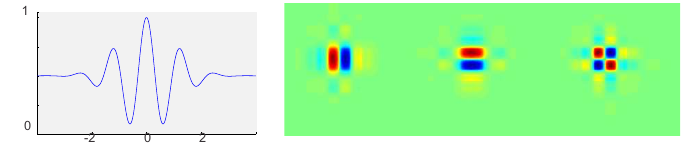}\\
  \end{center}
  \caption{One-dimensional wavelet template known as the Morlet wavelet (left)
and a two-dimensional wavelet displaying three different orientations:
vertical, horizontal and diagonal (right).}
\label{Fig1}
\end{figure}

\begin{figure}
  \begin{center}
  \includegraphics[
  width=12cm
  ]{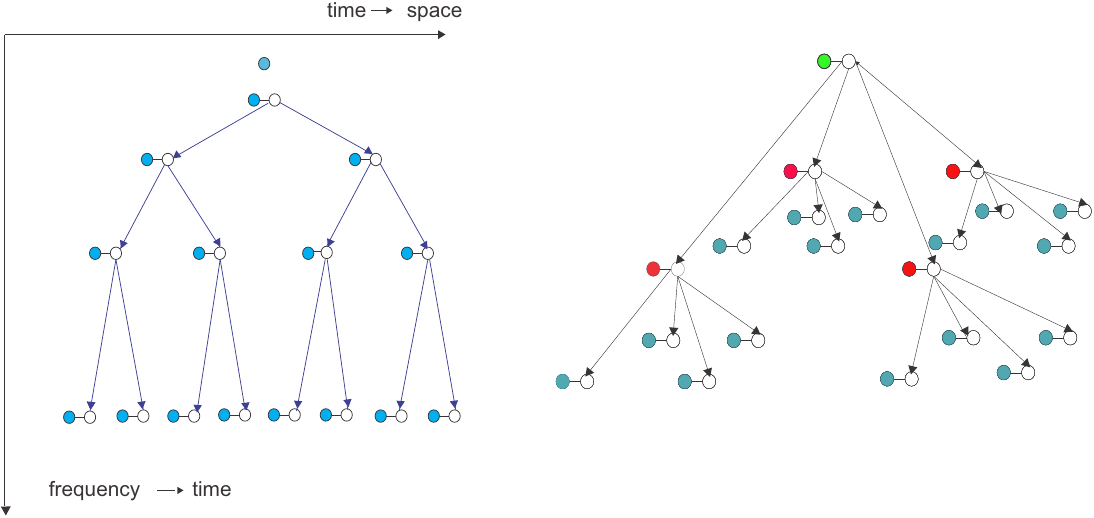}\\
  \end{center}
  \caption{Left: Statistical model of the one-dimensional wavelet transform. Each coefficient
(coloured node) is modeled as a mixture with the hidden state variable
(white node). The standard domain of the wavelet transform is time
(horizontal axis) - frequency (vertical axis) which in our model transforms
to the space-time domain. Hidden states are linked to each other vertically across
scales to yield the Hidden Markov tree. Right: Statistical model of the
two-dimensional wavelet tree (quad-tree). Nodes of the same colour belong to
the same scale.}
\label{Fig2}
\end{figure}

Since the interdependence (causal relationship) of the nodes takes place vertically
through the tree according to persistence property \cite{hern},\ we consider
time axis as directed from the coarsest to the finest scale although in a
conventional approach this axis represents frequency or scale. The domain of
the one-dimensional temporal signal is considered as spatial (intrinsic for
images) so that by introducing \textit{diffeomorphism invariance }the
wavelet tree becomes the spatio-temporal tree. The wavelet decomposition is
sparse implying that the number of large coefficients is small and the
number of small coefficients is large.\ The large coefficients, which we
call yang, convey information on singularities (1-D case) or edges (2-D
case) and the small, yin coefficients, contain information on smooth parts
of the signal or the image. The majority of the image energy is contained in
the yang coefficients, although the yin coefficients also store significant
energy, just because there are many of them. Usually the energy of the yin
coefficients is only one order lower than the total energy of the yang
coefficients while sometimes it may even surpass the yang energy. Thus, the
yin and yang coefficients of the wavelet decomposition are in a kind of
dynamic balance, justifying our choice of terminology. As a consequence of
the wavelet decomposition each coefficient has an associated probability
distribution indicating its frequency of occurrence. Usually, the
probability distributions, which are unknown ("hidden"), are modelled with
two zero-mean Gaussian distributions whose mixture is sufficient to model
the overall non-Gaussian distribution of wavelet coefficients. To each of
the two distributions corresponds a more frequently occurring (yin) state or
a less frequently occurring (yang) state. The locations in the image
containing sharp edges correspond to the less frequent, but more energy
containing yang coefficients, thus having a wider distribution at every
scale. The locations with prevailing smooth features correspond to the
narrow distribution since the corresponding yin coefficients are more
frequent although less energy containing. The corresponding hidden states
\textit{S} are labelled as 1 and 2, respectively (Fig.3).

\begin{figure}
  \begin{center}
  \includegraphics[
  width=6cm
  ]{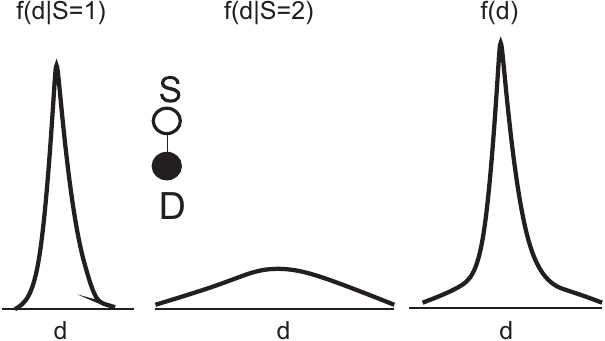}\\
  \end{center}
  \caption{Two-state, zero-mean Gaussian mixture model for wavelet
coeffiicients. Each wavelet coefficient is modeled with a hidden state
varaible S and a random variable D. The Gaussian conditional pdf's for a
low-variance state 1 (left) and a high-variance state 2 (middle) and the
overall non-Gaussian pdf (right) are shown.}
\label{Fig3}
\end{figure}

In simple terms, the hidden state of the 2-state model associated with each coefficient shows
whether the template wavelet overlaps an edge or not. Naturally, it is
possible to allocate probability distributions to a larger number of states,
if required. These states are modelled by the so called Hidden Markov Tree
Model (HMTM), characterized by the matrix whose entries are probabilities of
transition from one state to another. The probabilities of the hidden states
along with the probabilities of transition from one state to another and the
variances of the two distributions for each scale and orientation represent
the parameters of the HMM which are jointly evaluated by the Expectation
Maximization (EM) algorithm given the observed values of the wavelet
coefficients. Two important properties of the wavelet coefficients are
persistence and clustering implying respectively that the large or small
values of the coefficients tend to propagate across scales (in the vertical
direction of the (quad) tree) and the adjacent coefficients (in the
horizontal direction) tend to share the same properties. Due to the
persistence property which determines hierarchical causal dependencies,\ we
chose the direction of persistency propagation as the time axis, although in
the conventional approach this is the frequency axis. At first glance, such
practice suggests that we consider causality in a very weak sense, implying
that the outcome consistently proceeds from the cause which completely
determines it. However, mathematical framework, briefly explained earlier in
nontechnical terms, clearly reveals a probabilistic aspect of causality.

In order to simplify the model a standard procedure known as \textit{tying}
within the scale is used, so that variance and transition parameters are the
same at each scale of the wavelet transform. Such procedure enables
application to a limited number of images (e.g. one or two) without the need
of a training set. Also, it makes the model less image specific since it
rules out an a priori assumption on existence of smooth regions or edges at
certain spatial locations. As shown in \cite{we}, the hidden states are
actually the causal states which are sufficient for prediction purposes,
where prediction refers to the discovery of structure in the signal or in
the image. In analogy with \cite{crutchalizi}, the local statistical
complexity is defined as the entropy of the local causal state and the
global complexity is evaluated as the entropy of the whole hidden
(quad-)tree formed by the hidden states. The local complexity has a specific
physical interpretation in the sense that it is higher if the distribution
of the hidden yang and yin states in the node of the wavelet tree is more
uniform. In that case, there is a higher probability of the yang coefficient
appearance based on the persistence property contained in the nodes at the
immediate neighbouring scales meaning that the information stored in them
will be preserved. It is important to stress that the yin and yang states
are statistics of the complete tree of the wavelet coefficients, so that
separation into the future and the past becomes irrelevant to our
interpretation of causality. In spite of idiosyncracy of this method with
respect to the treatment of the past and the future, a similar conceptual
framework appeared already in physics. Namely, in the Feynman-Wheeler
picture of classical electrodynamics the radiation reaction of an
electrically charged particle is considered as an interaction with other
particles in both the past and the future \cite{feynman1} \cite{feynman2}.
In contrast to the conventional approach where the future action of the
particle may be determined by conditions at the present moment, in the
Feynman-Wheeler electrodynamics the future behaviour of the particles cannot
be predicted by specifying initial positions and velocities, but additional
information on the past and future behavior of the particles is required.

One of the crucial aspects of any wavelet based signal or image processing
technique is the choice of the optimal template (wavelet basis) so that
according to a certain predefined criterion it optimally corresponds to the
image. Our choice of the optimal wavelet basis is the one which maximizes
global statistical complexity. This criterion has been very successful in
determining and predicting properties of dynamical systems through the
analysis of times series \cite{we}, \cite{we2}, and here we extend the
application of this criterion in the context of art works where the
relationship between complexity and self-organization occurs naturally.
Hence, at the same time the method determines the optimal wavelet for each
particular image which, in turn, recognizes self-organization as a process
which increases local complexity in time evaluated as the maximal length of
the interval at which the complexity function increases monotonically. An
additional, special feature of this method is that it may be concurrently
used for noise reduction based on excellent denoising properties of wavelet
based HMM \cite{crouse}.

\section{Complexity, cognitive neuroscience and visual art}

As mentioned earlier, the fundamental course of our approach is based on the
importance of prediction and the information required for optimal
prediction. In order to gain deeper understanding of the basic ideas and
direction of our approach, it is significant to supplement it and contrast
it to the recently proposed predictive coding model of perception, an
important new direction of research in cognitive neuroscience \cite{milner},
\cite{friston1}, \cite{friston2}. According to this model the brain does not
passively register sensory input to which it is subjected, but actively
participates by making predictions \ based on experience. At every level of
visual hierarchy, which encompasses cortical structures of varying
complexity, predictions are made and propagated to lower levels (top-down)
where they are compared to the representation in the subordinate, lower
levels. The signals from the lower levels propagate in the opposite
direction (bottom-up). This comparison generates a prediction discrepancy or
prediction error which propagates to higher cortical levels where it
regulates the neuronal representation of sensory input and changes the
prediction. This self-organizing process takes place until the prediction
error is minimized leading to the generation of the most likely causal
input. It should be stressed that the prediction here refers to the
prediction of sensory effects from their cause and not the prediction of
sensory states in the future, i.e. forecasting. Each level in the cortical
hierarchy has a twofold function. First, it enables prediction based on the
information obtained from the lower level and second, it encodes the
mismatch between the generated prediction and the bottom-up evidence (the
prediction error) which is propagated to the next higher level of the
cortical hierarchy where further reduction of the prediction error takes
place. This hierarchical model is characterized by transfer of empirical
priors or constraints on the lower levels by the higher ones, thus it is
often attributed as the Bayesian brain model. As a result of this
hierarchical cortical process, the visual system organizes the perceptual
input in patterns, thus defining a structure which enables predictability of
visual representations. Reduction of prediction error, equal to the
free-energy in the model of Friston \cite{friston2}, arises from the
tendency of the brain and the whole body to retain homeostasis, an
equilibrium state. However, a perception of the work of art is significantly
different from the perception of ordinary things and events. Namely, art
requires complete involvement, which transcends simple observation so that
the observer acquires an active role as an accomplice or as a contender. The
role of emotions is also very important since in the ordinary perception the
mismatch between expectations and reality usually arouses negative emotion.
It is undeniable that repeated presentations in the works of art cause more
fluent and economical cortical processing due to increased predictability
(reduced prediction error), however that does not automatically imply
positive emotional arousal. An active observer, and particularly an art
connoisseur, expects to depart the default brain mode of preferred
predictability when observing the work of art and expects a reward, a
gratification in the form of a resolution of the prediction error which
results in a pleasurable aesthetic experience. Thus, an unpredictable visual
representation, causing a short-lived prediction error can be very effective
in causing pleasurable aesthetic experience, while redundancy of predictive
patterns may be boring and unemotional. It is not surprising that very often
artists, intuitively and sometimes precisely and according to a strict plan,
combine both predictable and unpredictable patterns in order to exert an
aesthetic impact. A good example of this practice are the works of M.C.
Escher, who induced prediction errors by combining repeated two-dimensional
patterns with optical illusions which suggest a higher-dimensional departure
from recurrent patterns. His artworks are also demonstrate how pleasurable
aesthetic experience may emanate even from a long-term predictive error.
Predictable perceptual forms and patterns may be periodically or
intermittently disconnected by patches that compel the viewer to complete
the visual experience, as practiced for example, by surrealist painters.
Predictable patterns may also be completely destroyed or fragmented in order
for new patterns and forms to appear. Resolution of prediction errors takes
place in the mind of the viewer and since paintings are static art forms it
induces dynamics which is stimulating and very often aesthetically pleasing.
It is not surprising that art viewers and appreciators expect prediction
errors and enjoy in resolving them while artists consciously create them and
sometimes use them as a kind of personal trade-mark. The interplay of
predictive patterns and unpredictable interruptions and the proportion of
their occurrence determines to a large degree the aesthetic experience and
gratification and has a strong impact on the emotional interpretation of the
work of art. An interesting view of the relationship between art and the
predictive coding model from the aspect of Gestalt psychology is given in
\cite{dercruys1}, \cite{dercruys2}.

There are three important features of the predictive coding model that
should be contrasted with our self-organization model. First, the predictive
coding model is a general model of perception which explains how the brain
retains its non-equilibrium steady state when subjected to visual stimuli.
Second, the resolution of the prediction error is necessary for the brain to
retain its steady state and the role of emotions may not be of importance in
completing the visual experience. Third, the resolution takes time and this
perceptual synthesis time or "time of contemplation" \cite{souriau}, is a
time experienced and created by the viewer. Our framework, which we may
refer to as the creativity model, is concerned with the work of art which
represents an authentic reflection of the self-organizing cognitive and
emotional processes taking place in the artist's brain. Thus, we indirectly
map the creative process of the artist's brain into a self-organizing
wavelet tree along with the statistical properties of the wavelet
coefficients. The brain of the artist in the act of creation is almost
without exception, not in equilibrium which leads to innovation and the
emergence of new ideas. Apparently the artist is trying to remain in such a
state until various possibilities of artistic expression are explored or
until the emergent ideas are actualized in the painting. The prediction
errors which may be manifested in the content, arrangement of forms,
aesthetic arrangements, colour juxtapositions, texture, design, etc., are
deliberately created in order to induce a specific aesthetic and emotional
impact on the observer and to induce the creative process of art
contemplation. However, the emotional and aesthetic feedback upon the artist
is also of great relevance, thus the artist creates and resolves the
predictive errors according to his mental and emotional state. Finally,
there is a specific form of time associated with the work of art,
essentially with its texture, best described by the term "the intrinsic time
of the work of art" \cite{souriau}, which we may be detected within our
framework. To quote Souriau, "There is no longer a question of a simple
psychological time of contemplation, but of an artistic time \textit{%
inherent in the texture itself of a picture} or a statue, in their
composition, in their aesthetic arrangement. Methodologically the
distinction is basic, and we come here (notably with Rodin's remark) to what
we must call the intrinsic time of the work of art. The significance of
these words (valid for any of the arts) is particularly clear and striking
when we deal with the representational arts, as in the normal case with
painting and sculpture (and also for literature, the theater, etc.)". The
psychological time mentioned by Souriau although mathematically intriguing
is beyond the current analysis and will be addressed elsewhere (M. R \& M.
M., in perparation). In our HMM wavelet model of self-organization, we
distinguish two different concepts of time. The first, referred to as the%
\textit{\ internal time} is recognized as the progression of causal
dependency among wavelet coefficients extending from the coarsest to the
finest scale. The internal time axis is graphically represented as the
vertical axis of the one-dimensional and the two-dimensional (quad-tree)
presented in Fig. 3. The second concept of time is related to the smooth
increase of local complexity and is completely determined by the
compositional and aesthetic arrangement of texture features of the image.
Since it coincides with the "intrinsic time of the work of art" of Souriau
we refer to it as the\textit{\ artistic time, }and it actually represents
one time frame of the internal time. In the next section this concept will
be illustrated and presented in more detail.

\section{Self-organization and complexity in the wavelet analysis of
paintings}

Although there are many measures of complexity, it is generally agreed that
the things which are completely random or completely uniform (or orderly)
are not complex. As a matter of fact, these two opposite aspects of the
disorder have zero complexity and the real complexity lies between them. The
maximal complexity corresponds to disorder lying somewhere close to halfway
between these two extrema. In the excerpt from \cite{arnheim}, presented
earlier, Arnheim mentions that the creation and communication of the
artistic idea is all mental; and we add here that the same is true of the
reception and understanding of this idea by the audience. Hence, the
artistic process creates a two-way information channel which contains
encoded and decoded symbols, where a symbol, in the context of paintings,
encompasses colours, texture, paints, brush strokes, forms, patterns, etc.
Based on the information exchanged in this channel, we find it appropriate
to adopt the concept and the first law of aesthetic complexity \cite{svozil}
which states that: "The aesthetics of artistic forms and designs depend on
their complexity. Too condensed coding makes a decryption of a work of art
impossible and is perceived as chaotic by the untrained mind, whereas too
regular structures are perceived as monotonous, too orderly and not very
stimulating". Accordingly, the more complex a pattern is in terms of
artistry and symbolisation, the more difficult is its decryption. A fast or
easy decryption may cause boredom while a difficult decryption may lead to
irritation and confusion. Hence the concept of aesthetic complexity may be
perceived as the general form of complexity. In our statistical complexity
approach we are focused on discovering causal relationships: how one symbol
leads to or brings about another symbol, thus establishing a direct
relationship between complexity, self-organization and creativity in art. In
order to illustrate our method we present the analysis results of two data
sets, one of which was previously analysed using different techniques and
which is freely available for download \cite{daubecheis}.

The first data set considered here consists of 7 high-resolution images of
paintings by the Dutch artist Charlotte Caspers. She was commissioned by
Ingrid Daubecheis and the members of the Machine Learning and Image
Processing for Art Investigation Research Group at Princeton University to
paint 7 paintings of relatively small size (approximately 25 cm x 20 cm) of
different styles and using different materials\cite{daubecheis}. Within the
next few days she has also painted a copy for each painting using the same
paints, brushes and grounds and under the same lighting conditions. For the
presentation of our method and the results of the analysis, it is of
interest to mention the remark of I. Daubecheis \cite{daubecheis} that C.
Caspers spent close to 2 times more time on creating each copy as compared
to the original, indicating that "painting a copy is a more painstaking
process than the spontaneous painting of an original". The copies were of
such high quality that the artist was convinced that it would not be able to
distinguish copies from originals. The high-resolution digital images were
downloaded from the home site of the Princeton group\footnote{%
http://web.math.princeton.edu/ipai/index.html}.

In the so called RGB (Red, Green, Blue) colour space, each pixel in a colour
image is represented by red, green and blue components. Each component may
be treated as a separate image and for each painting we perform the analysis
for each colour separately. The wavelet transform was applied in a twofold
manner, namely on the whole painting and on all the patches of size 512 x
512 pixels, applying overlapping where necessary due to the dimension of the
painting. The top nine scales of the transformation are used to form the
causal structure which represents the cornerstone of our method. Note that
all of the methods for artist authentication based on the wavelet
decomposition apply the opposite practice, namely they use the coefficients
at the few finest scales which contain the majority of coefficients. The
templates from the standard orthogonal and biorthogonal wavelet families are
used: Haar (haar), Daubechies (db2), Symlet (sym3), Coiflet (coif1),
Biorthogonal (bior1.3), Reverse Biorthogonal (rbior1.3) and Discrete Meyer
(dmey). The wavelet transform has a layered structure, where each layer
corresponds to a particular scale. Each layer consists of the slightly
blurrier version of the image and the wavelet transforms along the three
directions (horizontal, vertical and diagonal) which need to be complemented
with detail information in order to reconstruct the original image.
Parameters of the HMM\ model are evaluated for two, three, four and five
hidden states, however, we have found no substantial difference between
results for the two-state and for the higher states, so that only
paradigmatic results for the two-state case are presented. The local
complexity is evaluated as the Shannon entropy of the hidden variables in
each node (coefficient) and the global entropy is evaluated as the entropy
of the whole wavelet tree. The crucial importance of the global entropy is
that it measures the increase of complexity in time so that
self-organization of various degrees may be recognized, for example weak and
strong self-organization may be defined accordingly. Higher global entropy
implies stronger self-organization requiring more information for prediction
while the weaker self-organization has the opposite attributes.
Few of the paintings from this set and their images are presented in Figs.
4, 5 and 6.

\begin{figure}
  \begin{center}
  \includegraphics[
  width=12cm
  ]{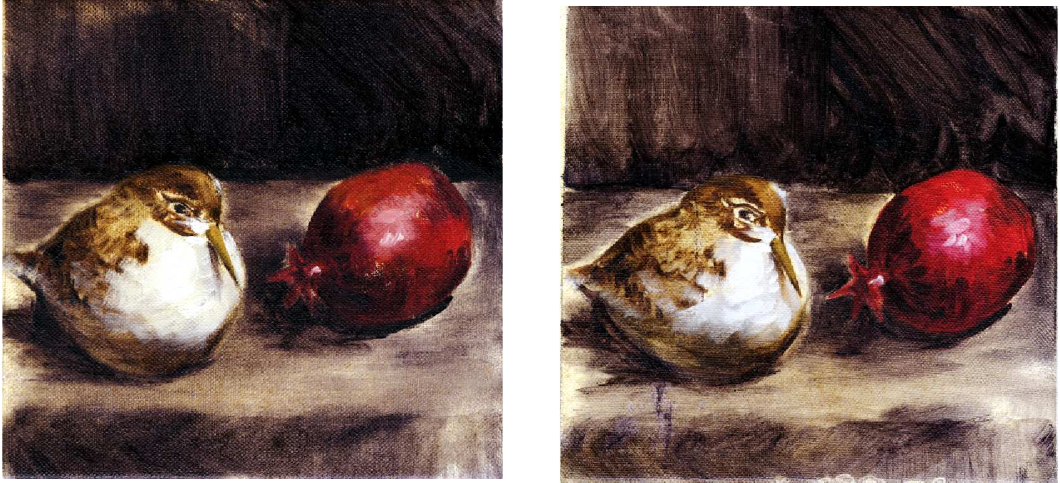}\\
  \end{center}
  \caption{A still-life painting by Charlotte Caspers commisioned by Ingrid Daubechies
and the Art Investigation group at Princeton University.\ The canvas is a
crude, absorbing jute-type. Soft brushes were used.}
\label{Fig4}
\end{figure}

\begin{figure}
  \begin{center}
  \includegraphics[
  width=12cm
  ]{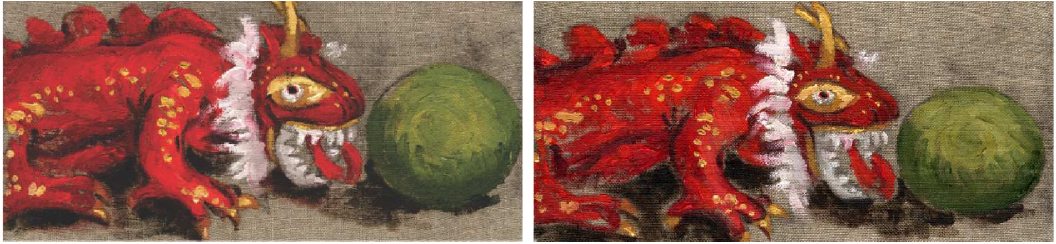}\\
  \end{center}
  \caption{The second set of paintings by Charlotte Caspers, based on the use of
acrylic paint.The brush strokes on a commercially primed canvas are visibly
accentuated. Both soft and hard brusheds were used.}
\label{Fig5}
\end{figure}

\begin{figure}
  \begin{center}
  \includegraphics[
  width=12cm
  ]{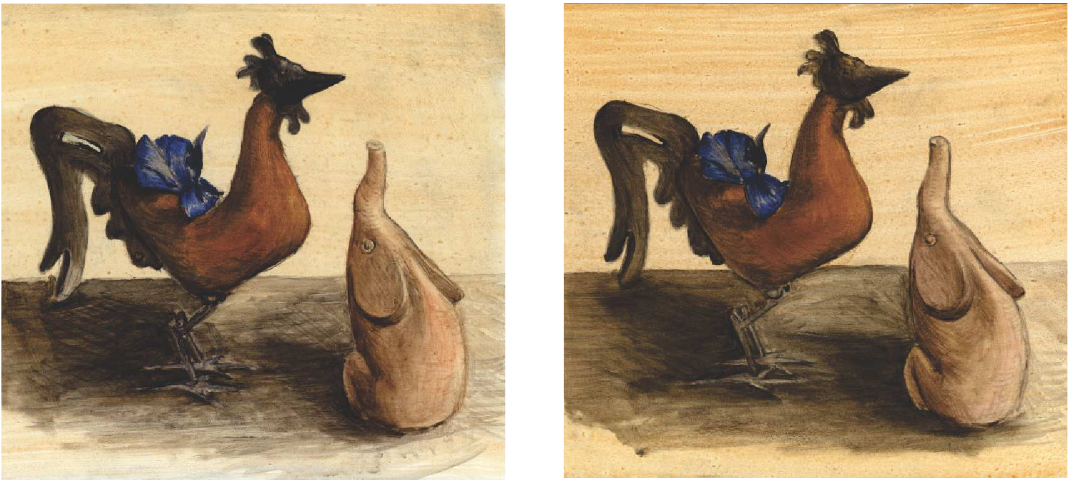}\\
  \end{center}
  \caption{The third set of paintings by Charlotte Caspers painted with oil paints and
soft brushes on a chalk-ground. The technique is similar to the 15-th
century Flemish paintings.}
  \label{Fig6}
\end{figure}

The two dimensional wavelet-transform acts as an orientation microscope,
which detects discontinuities of images such as point singularities (contour
vertices) or orientation features such as edges, borders, segments or
interwoven, mikado type edges. In Fig 7 we present local complexity of the
painting and its copy presented in Fig 4 as an illustration of some of the
representative characteristics of this quantity. Continual, smooth increase

\begin{figure}
  \begin{center}
  \includegraphics[
  width=12cm
  ]{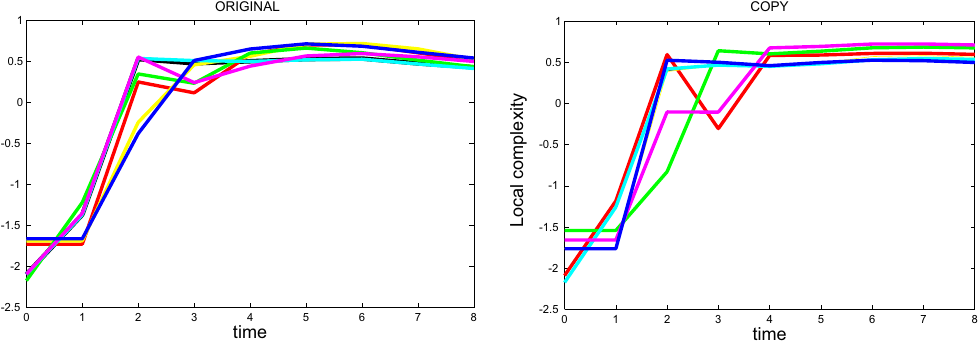}\\
  \end{center}
  \caption{Local complexity of the paintings shown in Fig. 4. Each color
corresponds to a different wavelet: black (Haar), red (db2), yellow (sym3),
green (coif1), magenta (bior1.3), purple (rbior1.3) and blue (dmey).}
  \label{Fig7}
\end{figure}

of local complexity indicates self-organization and the corresponding time
during which the self-organization takes place we refer to as the time of
self-organization. In most of the cases considered, local complexity
corresponding to the original painting has smoother characteristics and
displays longer self-organization for the majority of wavelets than in the
case of a copy. For example, the rbior1.3 wavelet displayed in blue colour,
exhibits the long-term self-organization which spans almost 5 time units in
the case of the original painting (left) while it displays self-organization
only during one time unit in the case of a copy (right). From the aspect of
local complexity the optimal wavelet is the one which displays the
lengthiest self-organization, time-wise. However, as mentioned earlier
global complexity quantifies self-organization for the whole wavelet tree
and represents the most important measure of complexity and self-organization.

It is useful to consider and compare global complexity characteristics in
different orientations, so the global complexity values are presented
corresponding to horizontal, vertical and diagonal directions. The mean
value of these orientational complexities is the most important quantity
which we simply refer to as the global complexity. First, we present typical
results of the analysis performed on the whole painting and Table 1 displays
orientational complexities and the global complexity for the red component
of the paintings presented in Fig. 4.

\begin{table}[h!]
\centering
\begin{tabular}{|l|l|l|l|l|l|l|l|}
\hline
\multicolumn{8}{|l|}{\textbf{Global complexity}; paintings Fig. 4, colour:
red} \\ \hline
Wavelet & haar & db2 & sym3 & coif1 & bior1.3 & rbior1.3 & dmey \\ \hline
Original & 0.7765 & 0.6258 & 0.6763 & 0.7435 & 0.7644 & 0.7921 & \textbf{%
0.8134} \\ \hline
Copy & 0.7804 & 0.7178 & 0.7496 & 0.6298 & 0.7841 & \textbf{0.7856} & 0.5674
\\ \hline
\multicolumn{8}{|l|}{\textbf{Diagonal complexity}} \\ \hline
Original & 0.8992 & 0.6500 & 0.7196 & 0.8945 & 0.8895 & 0.9492 & \textbf{%
0.9534} \\ \hline
Copy & 0.8977 & 0.8052 & 0.8170 & 0.8933 & 0.8922 & \textbf{0.9034} & 0.7852
\\ \hline
\multicolumn{8}{|l|}{\textbf{Horizontal complexity}} \\ \hline
Original & 0.7845 & 0.7222 & 0.8130 & 0.7754 & 0.7486 & 0.7832 & \textbf{%
0.8278} \\ \hline
Copy & 0.7896 & 0.8365 & \textbf{0.8188} & 0.8045 & 0.8044 & 0.7342 & 0.8026
\\ \hline
\multicolumn{8}{|l|}{\textbf{Vertical complexity}} \\ \hline
Original & 0.6459 & 0.5053 & 0.4962 & 0.5607 & 0.6552 & 0.6439 & \textbf{%
0.6591} \\ \hline
Copy & 0.6537 & 0.5117 & 0.6129 & 0.4191 & \textbf{0.6450} & 0.6191 & 0.6149
\\ \hline
\end{tabular}
\caption{The global complexity and the orientational complexities evaluated for the
entire paintings of Fig. 4. The colour in the RGB colour space is red. The optimal wavelet
is denoted in bold.}
\label{table:1}
\end{table}
\bigskip
The optimal wavelet corresponds to the maximum global complexity and is
marked in bold. In Table 2. the global and directional complexities are
presented for the red component of the paintings presented in Fig. 5.
Similar results are obtained for the green and blue components and are not shown here.
\bigskip

\begin{table}[h!]
\centering
\begin{tabular}{|l|l|l|l|l|l|l|l|}
\hline
\multicolumn{8}{|l|}{\textbf{Global complexity}; painting Fig. 4, colour: green
} \\ \hline
Wavelet & haar & db2 & sym3 & coif1 & bior1.3 & rbior1.3 & dmey \\ \hline
Original & \textbf{0.4515} & 0.4284 & 0.4368 & 0.4240 & 0.4438 & 0.4339 &
0.3875 \\ \hline
Copy & 0.4177 & 0.4453 & 0.3936 & 0.4027 & \textbf{0.4244} & 0.3954 & 0.3890
\\ \hline
\multicolumn{8}{|l|}{\textbf{Diagonal complexity}} \\ \hline
Original & 0.5123 & 0.4601 & 0.4594 & 0.4445 & \textbf{0.5151} & 0.4306 &
0.4314 \\ \hline
Copy & 0.4753 & 0.46163 & 0.41101 & 0.4073 & \textbf{0.4868} & 0.4091 &
0.4242 \\ \hline
\multicolumn{8}{|l|}{\textbf{Horizontal complexity}} \\ \hline
Original & \textbf{0.4179} & 0.3945 & 0.3988 & 0.3921 & 0.3976 & 0.4147 &
0.3368 \\ \hline
Copy & 0.3759 & \textbf{0.4171} & 0.3175 & 0.3554 & 0.3400 & 0.3342 & 0.3279
\\ \hline
\multicolumn{8}{|l|}{\textbf{Vertical complexity}} \\ \hline
Original & 0.4242 & 0.4303 & 0.4523 & 0.4355 & 0.4168 & \textbf{0.4566} &
0.3942 \\ \hline
Copy & 0.4019 & \textbf{0.4551} & 0.4521 & 0.4452 & 0.4464 & 0.4428 & 0.4148
\\ \hline
\end{tabular}
\caption{The global complexity and the orientational complexities evaluated for the
entire paintings of Fig. 5. The colour in the RGB colour space is red. The optimal wavelet
is denoted in bold.}
\label{table:2}
\end{table}
\bigskip

These Tables capture paradigmatic characteristics of all paintings from this
set with respect to self-organization and complexity. First, the global
complexity corresponding to the optimal wavelet of original paintings is
always larger than the reciprocal global complexity of copies. Second, the
optimal wavelet may be the same for all orientations, including the global
self-organization indicator although this is not the rule. It is natural to
agree that the complexity of the painting may be different in different
directions depending on, for example, artist's technique or the thematic
content. For example, in Table 1 the discrete Meyer wavelet persists as an
optimal wavelet for the original painting for all directions as well as for
the global complexity, while the optimal wavelet for the copy is the same
(rbior1.3) only for the diagonal direction and for the global complexity.
This inconsistency is also an important indicator of the lack of
self-regulating flow of ideas which materializes on canvas, or it could be
the result of a long creative process which takes days to complete and which
lacks creative continuity. However, from the information provided in ref.%
\cite{daubecheis} we may rule out the latter instance. We have found that
for almost all the paintings the maximum global complexity and the maximum
diagonal complexity correspond to the same (optimal) wavelet, or, in rare
cases when this condition is not fulfilled the value of the diagonal
complexity corresponding to the optimal wavelet is very close to the maximal
diagonal complexity.\ For example, in Table 2 the maximum diagonal
complexity corresponds to the rbior1.3 wavelet, however the value
correlative to the optimal Haar wavelet is very close to the value of the
rbior1.3 wavelet.

The second approach is based on the analysis of patches of size 512 x 512
pixels, and in general we have found that the distinction between originals
and copies is more apparent than in the case when the whole painting is
subject to the wavelet HMM analysis. Tables 3 and 4 present global
complexity results for one of the patches of paintings in Fig. 4 and 5
respectively.

\begin{table}[h!]
\centering
\begin{tabular}{|l|l|l|l|l|l|l|l|}
\hline
\multicolumn{8}{|l|}{\textbf{Global complexity}; paintings Fig. 4, patch 512
x 512, colour: red} \\ \hline
Wavelet & haar & db2 & sym3 & coif1 & bior1.3 & rbior1.3 & dmey \\ \hline
Original & 0.5470 & 0.5508 & 0.5543 & \textbf{0.5648} & 0.5187 & 0.5332 &
0.4831 \\ \hline
Copy & 0.5116 & 0.5382 & 0.5091 & \textbf{0.5326} & 0.5010 & 0.5049 & 0.4236
\\ \hline
\end{tabular}
\caption{The global complexity evaluated for one patch of the size 512 x 512 pixels
of the paintings shown in Fig. 4. The colour of the RGB spectrum is red.}
\label{table:3}
\end{table}
\bigskip

\begin{table}[h!]
\centering
\begin{tabular}{|l|l|l|l|l|l|l|l|}
\hline
\multicolumn{8}{|l|}{\textbf{Global complexity}; paintings Fig. 6, patch 512
x 512, colour: red} \\ \hline
Wavelet & haar & db2 & sym3 & coif1 & bior1.3 & rbior1.3 & dmey \\ \hline
Original & 0.5772 & \textbf{0.5900} & 0.5807 & 0.5797 & 0.5775 & 0.5696 &
0.4878 \\ \hline
Copy & 0.2686 & 0.2883 & 0.2967 & 0.2931 & 0.2610 & \textbf{0.3079} & 0.2832
\\ \hline
\end{tabular}
\caption{The global complexity evaluated for one patch of the size 512 x 512 pixels
of the paintings shown in Fig. 5. The colour of the RGB spectrum is red.}
\label{table:4}
\end{table}

For all patches and all the paintings from the set the mean global
complexity of an original painting is larger than the corresponding value of
a copy. We have found that in a few cases the horizontal or the vertical
complexity of a copy may be larger than the analogous value of an original,
however the contributions of other two orientational complexities prevail and the mean
global complexity is always larger for an original.

The second set of images consists of two paintings by Rene Magritte, known
under the title "La saveur des larmes" ("The flavour of tears"). The
paintings are presented side by side in\ Fig. 8. One is an original, but
which one? And which one is a copy? One is in the Barber Museum of Fine Arts in Birmingham, UK and the other in
the Mus\'{e}es Royaux des Beaux Arts de Belgique in Brussels. The canvases
are both dated 1948 and since Rene Magritte was a Surrealist with an
exquisite sense of humour he might have been enjoying a charming and
probably profitable joke. Magritte may well have seen his forgeries as part
of the conflict between the real and the unreal, as the tension between
these two realms was one of the hallmarks of the Surrealist movement.
Magritte is known to have played a joke with the audience when he hung his
forgery of Max Ernst's painting "The Forest" in place of the original in
1943. Giorgio de Chirico, another famous surrealist, in his later years
created what he called "self-forgeries" of the paintings from his earlier
period. He would backdate them with an intention to make fun of the art
critics as a revenge for their critique of his later works.

\begin{figure}
  \begin{center}
  \includegraphics[
  width=12cm
  ]{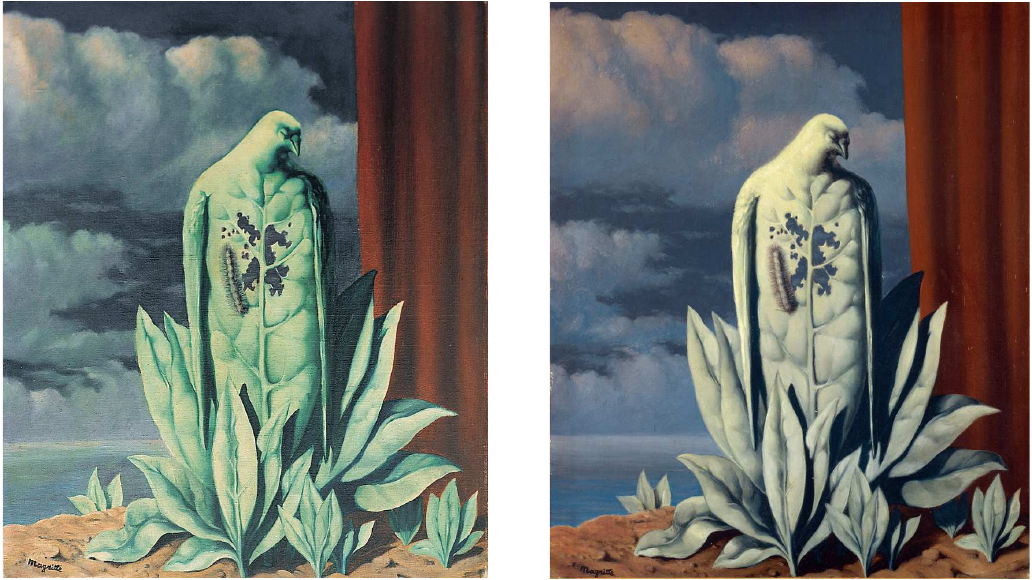}\\
  \end{center}
  \caption{Two paintings "The Flavour of tears" by Ren\'{e} Magritte. Which one
is an original and which one is a copy?}
\label{Fig8}
\end{figure}

Art experts now consider that both canvases are Magritte originals, and
assume he forged his own work to make money during the war years. The
existence of both paintings was unknown until 1983 when one of the canvases
turned up at an auction in New York while the other remained in Europe. The
two versions of the same painting are identical by all means and the experts
agree that even the holes made by a caterpillar are exactly the same on the
two canvases. Even the inscriptions on the back of the paintings are the
same and undiscernible. So, which one of the two canvases may be considered
as an original? We show that indisputably one of them has more indicators of
creative artistic idea transferred on canvas, then the other so we claim
with utmost confidence, that only one of them is the result of
self-regulatory creative work. The other is a copy by the original artist.
In order to distinguish them in the text we refer to them as "The flavour of tears 1" and "The flavour of tears 2".
As an illustration, in Tables 5 and 6 we present a comparison of global and
orientational complexities of the two paintings for the dominant colours of the RGB spectrum, namely the blue and
the green colour respectively. Similar results leading to the same conclusions are obtained for the red colour,
and are not presented here. The analysis is preformed on the entire painting.

\begin{table}[h!]
\centering
\begin{tabular}{|l|l|l|l|l|l|l|l|}
\hline
\multicolumn{8}{|l|}{\textbf{Global complexity}; "The Flavour of Tears 1 and
2"; colour: blue} \\ \hline
Wavelet & haar & db2 & sym3 & coif1 & bior1.3 & rbior1.3 & dmey \\ \hline
1 & 0.1845 & 0.2804 & \textbf{0.3186} & 0.2612 & 0.1950 & 0.2225 & 0.2780 \\
\hline
2 & 0.1795 & 0.2427 & \textbf{0.2905} & 0.2807 & 0.1918 & 0.1979 & 0.2711 \\
\hline
\multicolumn{8}{|l|}{\textbf{Diagonal complexity}} \\ \hline
1 & 0.1849 & 0.4275 & \textbf{0.5211} & 0.3706 & 0.1824 & 0.2241 & 0.3917 \\
\hline
2 & 0.1668 & 0.3201 & \textbf{0.3707} & 0.2778 & 0.1682 & 0.1581 & 0.3448 \\
\hline
\multicolumn{8}{|l|}{\textbf{Horizontal complexity}} \\ \hline
1 & 0.2010 & 0.2048 & 0.2088 & 0.2066 & 0.2018 & \textbf{0.2293} & 0.2087 \\
\hline
2 & 0.2138 & 0.2247 & 0.2179 & \textbf{0.2250} & 0.2123 & 0.2113 & 0.2016 \\
\hline
\multicolumn{8}{|l|}{\textbf{Vertical complexity}} \\ \hline
1 & 0.1677 & 0.2088 & 0.2298 & 0.2063 & 0.1708 & 0.2141 & \textbf{0.2335} \\
\hline
2 & 0.1578 & 0.1831 & \textbf{0.1878} & 0.1760 & 0.1596 & 0.1755 & 0.1688 \\
\hline
\end{tabular}
\caption{The global complexity and the orientational complexities evaluated for the
Magritte's paintings. The colour in the RGB colour space is blue. The optimal wavelet
is denoted in bold.}
\label{table:5}
\end{table}
\bigskip
\bigskip
\begin{table}[h!]
\centering
\begin{tabular}{|l|l|l|l|l|l|l|l|}
\hline
\multicolumn{8}{|l|}{\textbf{Global complexity}; "The Flavour of Tears 1 and
2"; colour: green} \\ \hline
Wavelet & haar & db2 & sym3 & coif1 & bior1.3 & rbior1.3 & dmey \\ \hline
1 & 0.1907 & 0.2836 & \textbf{0.3209} & 0.2663 & 0.1908 & 0.2264 & 0.2840 \\
\hline
2 & 0.1746 & 0.2396 & \textbf{0.2560} & 0.2251 & 0.1747 & 0.1802 & 0.2442 \\
\hline
\multicolumn{8}{|l|}{\textbf{Diagonal complexity}} \\ \hline
1 & 0.2013 & 0.4322 & \textbf{0.5219} & 0.3816 & 0.1987 & 0.2367 & 0.4049 \\
\hline
2 & 0.1620 & 0.3195 & \textbf{0.3691} & 0.2812 & 0.1637 & 0.1556 & 0.3585 \\
\hline
\multicolumn{8}{|l|}{\textbf{Horizontal complexity}} \\ \hline
1 & 0.2085 & 0.2074 & \textbf{0.2271} & 0.2192 & 0.1994 & 0.2265 & 0.2126 \\
\hline
2 & 0.2113 & 0.2222 & \textbf{0.2154} & 0.2145 & 0.2009 & 0.2089 & 0.2043 \\
\hline
\multicolumn{8}{|l|}{\textbf{Vertical complexity}} \\ \hline
1 & 0.1722 & 0.2114 & 0.2237 & 0.2080 & 0.1741 & 0.2160 & \textbf{0.2346} \\
\hline
2 & 0.1504 & 0.1771 & \textbf{0.1834} & 0.1697 & 0.1515 & 0.1762 & 0.1699 \\
\hline
\end{tabular}
\caption{The global complexity and the orientational complexities evaluated for the
Magritte's paintings. The colour in the RGB colour space is green. The optimal wavelet
is denoted in bold.}
\label{table:6}
\end{table}

It can be immediately noticed that the Symlet (sym3) is the optimal wavelet
for both colour cases and that the global complexity and the diagonal
complexity for the painting 1 are considerably larger than for the painting
2. Also, the optimal wavelet values for horizontal and vertical directions
are also in favour of painting 1 so that we may ascribe to this canvas the
attribute of originality in the sense that it represents an outcome of the
creative artistic expression. Similar results are obtained when the analysis
is performed on patches of size 512 x 512 pixels. Tables 7 and 8 illustrate
typical results of this analysis.

\begin{table}[h!]
\centering
\begin{tabular}{|l|l|l|l|l|l|l|l|}
\hline
\multicolumn{8}{|l|}{\textbf{Global complexity}; "The Flavour of tears 1 and
2", patch A, colour: blue} \\ \hline
Wavelet & haar & db2 & sym3 & coif1 & bior1.3 & rbior1.3 & dmey \\ \hline
1 & 0.1888 & 0.2754 & \textbf{0.3231} & 0.2564 & 0.1882 & 0.2176 & 0.2717 \\
\hline
2 & 0.1712 & 0.2410 & \textbf{0.2545} & 0.2216 & 0.1721 & 0.1800 & 0.2378 \\
\hline
\end{tabular}
\caption{The global complexity evaluated for one, typical patch of the size 512 x 512 pixels
of the Magritte's paintings. The colour of the RGB spectrum is blue.}
\label{table:7}
\end{table}
\bigskip

\begin{table}
\centering
\begin{tabular}{|l|l|l|l|l|l|l|l|}
\hline
\multicolumn{8}{|l|}{\textbf{Global complexity}; "The Flavour of tears 1 and
2", patch B, colour: green} \\ \hline
Wavelet & haar & db2 & sym3 & coif1 & bior1.3 & rbior1.3 & dmey \\ \hline
1 & 0.1907 & 0.2836 & \textbf{0.3209} & 0.2665 & 0.1906 & 0.2276 & 0.2840 \\
\hline
2 & 0.1888 & 0.2754 & \textbf{0.3131} & 0.2566 & 0.1882 & 0.2176 & 0.2718 \\
\hline
\end{tabular}
\caption{The global complexity evaluated for one, typical patch of the size 512 x 512 pixels
of the Magritte's paintings. The colour of the RGB spectrum is green.}
\label{table:8}
\end{table}
\bigskip

An interesting feature of these\ results is that they show remarkable
consistency between self-organisation indices for the entire painting and for
patches of size 512 x 512 pixels when either of the two dominant colours,
blue and green, are analysed. Similar results are obtained for the red
colour (not shown) and for patches positioned in the areas dominated by the
red colour (drapes on the right hand side of the paintings). Minor
departures from this trend were noticed in a small number of cases, and only
for horizontal or vertical complexities. The consistency of the results
suggests that Magritte was highly skilled in copying his own work and that
perhaps he devised a special technique for that purpose, a practice that
would be in the spirit of surrealism and surrealists.

\section{Conclusion}
Several writers on art as either theoreticians or practitioners, or both,
have rightfully hypothesized that artistic creativity has many properties in
accord with nonlinear dynamics, e.g. in \cite{zausner} and references
therein. However, even before the advent of chaos theory and non-equilibrium
thermodynamics, it was suggested that self-regulatory and self-organizing
processes may be recognized in the works of art. The mind of an artist is an
open, dissipative system which absorbs information from the external world
and produces entropy which could take the form of an artwork. We suggest
here that artistic creativity, generally perceived as aesthetic or pleasing
is self-organisation process which could be detected by examining the work
of art and we focus our attention on paintings which, unlike other forms of
art, are frequently subject to forgeries. When the work of art is created
different elements, forms and textures are assembled and juxtaposed in a
specific way, creating a higher organization than in the case when the same
constituents are by themselves. Self-organizing processes in the brain of an
artist create ideas and emotions which, by means of the artist's brush
stroks are transferred on canvas creating "higher organization of meaning in
the work of art". We show that complexity and self-organisation are
numerical quantities which could be used to differentiate between an
original, creative, artistic intension and realization from the technical
process which produces a copy of the work of art. Although the method shows
very good and promising results in recognizing creative process, further
improvements are possible. A non-exhaustive list of some of the advantages
of the presented framework in comparison with the existing methods for art
work analysis and artist authentication are the following:

In order to obtain reliable and conclusive results it is not necessary to
use the complex wavelet transform which provides three additional
orientations with respect to the ordinary wavelet transform.

A sophisticated analysis of brush strokes is not necessary and it is not
necessary to use separate statistical models for texture-based and
brush-stroke features.

It is not necessary to know which work of art is original as long there is
another art work for comparison from the aspect of originality and
creativity.

It is not necessary to have a training set consisting of other works of art
of the same artist.

It is not necessary to introduce various distance (dissimilarity) measures

Finally, we regard our work as an important step in integration of artistic
and scientific perspectives and an attempt in creating common ground for
communication of ideas between arts and sciences.

\begin{acknowledgement}
The authors thank the Societ\'{e} des auteurs dans les arts graphiques et
plastiques (ADAGP), Paris for sending us the high quality digital copies of
Magritte's works and for permission to use the reproductions in our
publications. The authors acknowledge the financial support by the Serbian
ministry of education, science and technology under grant OI174014.
\end{acknowledgement}

\end{document}